%% file: main.tex
\documentclass[10pt,twocolumn,letterpaper]{article}

\usepackage[pagenumbers]{cvpr} % To force page numbers, e.g. for an arXiv version

\pdfoutput=1

\input{preamble}

\definecolor{cvprblue}{rgb}{0.21,0.49,0.74}
\usepackage[pagebackref,breaklinks,colorlinks,citecolor=cvprblue]{hyperref}

\def\paperID{16544} % *** Enter the Paper ID here
\def\confName{CVPR}
\def\confYear{2024}

\title{Unsupervised Video Domain Adaptation with Masked Pre-Training and Collaborative Self-Training}

\author{Arun Reddy$^{1,2}$, William Paul$^{1}$, Corban Rivera$^{1}$, Ketul Shah$^{2}$, Celso M. de Melo$^{3}$, Rama Chellappa$^{2}$\\
\small ${}^1$Johns Hopkins University Applied Physics Laboratory \\
\small ${}^2$Johns Hopkins University, Department of Electrical \& Computer Engineering \\
\small ${}^3$DEVCOM U.S. Army Research Laboratory\\
}

\begin{document}
\maketitle
\input{sec/0_abstract}    
\input{sec/1_intro}

\input{sec/2_related}

\input{sec/3_preliminaries}
\input{sec/4_method}
\input{sec/5_experiments}

\input{sec/6_analysis}

\input{sec/7_conclusions}
\clearpage
{
    \small
    \bibliographystyle{ieeenat_fullname}
    \bibliography{main}
}

\input{sec/X_suppl}

\end{document}

%% file: preamble.tex
\usepackage[accsupp]{axessibility} % Improves PDF readability for those with visual impairments.
\usepackage[dvipsnames]{xcolor}
\usepackage{multirow}
\usepackage{colortbl}

\newcommand{\methodname}{UNITE\xspace}

\definecolor{celadon}{rgb}{0.67, 0.88, 0.69}

\newcommand{\ourcolor}{\cellcolor{celadon!20}}
\newcommand{\socolor}{\cellcolor{Melon!15}}
\newcommand{\tocolor}{\cellcolor{Gray!15}}

\usepackage{pifont}% http://ctan.org/pkg/pifont
\newcommand{\cmark}{\ding{51}}%
\newcommand{\xmark}{\ding{55}}

\usepackage{fancyhdr}

%% file: sec/0_abstract.tex
\begin{abstract}
In this work, we tackle the problem of unsupervised domain adaptation (UDA) for video action recognition. Our approach, which we call UNITE, uses an image teacher model to adapt a video student model to the target domain. UNITE first employs self-supervised pre-training to promote discriminative feature learning on target domain videos using a teacher-guided masked distillation objective. We then perform self-training on masked target data, using the video student model and image teacher model together to generate improved pseudolabels for unlabeled target videos. Our self-training process successfully leverages the strengths of both models to achieve strong transfer performance across domains. We evaluate our approach on multiple video domain adaptation benchmarks and observe significant improvements upon previously reported results. Code is available at \href{https://github.com/reddyav1/unite}{\texttt{https://github.com/reddyav1/unite}}.
\end{abstract}

%% file: sec/1_intro.tex
% Define a custom page style for the first page
\fancypagestyle{firstpage}{
  \fancyhf{} % Clear header and footer
  \renewcommand{\headrulewidth}{0pt} % Remove header rule
  \lfoot{} % Left footer content
  \cfoot{\thepage} % Center footer content (leave empty)
  \rfoot{\textit{Approved for public release. Distribution is unlimited.}} % Right footer content (leave empty)
}
% \fancyfoot{} % clear all footer fields

\thispagestyle{firstpage}

\section{Introduction}
\label{sec:intro}

% Outline
% - Introduce video action recognition
% - Describe domain shift (performance degradation caused by differences in data distribution during training and test) and need for domain adaptation techiques 
% - Describe why this matters for video recognition
% - If we decide to focus on video datasets with large gaps (HMDB-ARID, Mixamo-Kinetics, RoCoG-v2), describe how existing benchmarks are very limited in how they assess techniques. Improved generalization vs. explicit adaptation?
% - Limitations of existing approaches and high-level motivations behind our approach
% - Describe our approach in general terms
% - Bulleted list of our paper's contributions

In recent years, the field of video action recognition has undergone significant advancement, largely driven by deep learning-based techniques. These include approaches based on convolutional neural networks (CNNs) \cite{simonyanTwoStreamConvolutionalNetworks2014, tranLearningSpatiotemporalFeatures2015, carreiraQuoVadisAction2017, wangTemporalSegmentNetworks2017, feichtenhoferSlowFastNetworksVideo2019, feichtenhoferX3DExpandingArchitectures2020} and vision transformers (ViTs) \cite{piergiovanniRethinkingVideoViTs2023, arnabViViTVideoVision2021, liuVideoSwinTransformer2022, fanMultiscaleVisionTransformers2021, bertasiusSpaceTimeAttentionAll2021, neimarkVideoTransformerNetwork2021}. Further enriching this landscape, models integrating video and language data have emerged \cite{wangActionCLIPNewParadigm2021, maXCLIPEndtoEndMultigrained2022, wangInternVideoGeneralVideo2022}, which effectively leverage large collections of captioned videos to achieve impressive video understanding capabilities.

Despite increasing performance on benchmark datasets, deep networks for video action recognition suffer from the same fundamental challenges as other machine learning models in the presence of \textit{distribution shift}. This phenomenon, where a model's performance degrades when applied to data distributions different from those it was trained on, remains a critical obstacle in deploying the models in varied real-world scenarios.

\begin{figure}[t]
  \centering
    \includegraphics[width=0.9\linewidth]{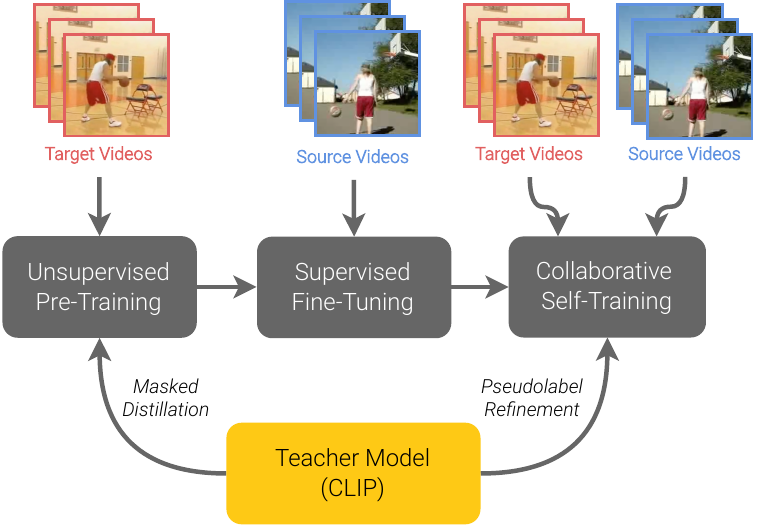}

   \caption{Overview of the \methodname pipeline for video UDA. A teacher model is used to guide the self-supervised learning process in the first stage, and is again used to improve pseudolabeling of target videos during self-training in the third stage.}
   \label{fig:teaser}
    \vspace{-10pt}
  \setlength{\abovecaptionskip}{-6pt}
  \setlength{\belowcaptionskip}{-6pt}
\end{figure}

Domain adaptation (DA) seeks to mitigate the effects of distribution shift and has been a major research area within the computer vision community for years \cite{patelVisualDomainAdaptation2015}. A subproblem within DA is unsupervised domain adaptation (UDA), which tries to make use of unlabeled data from the target domain to improve transfer performance. UDA approaches specific to video have also been developed, with many focusing on explicit alignment of the source and target domains using adversarial techniques or domain discrepancy measures. In contrast, we present here a method for video UDA that relies on self-supervised masked pre-training and self-training, both aided by a powerful image-based teacher model (CLIP \cite{radfordLearningTransferableVisual2021a} ). The authors of DALL-V \cite{zaraUnreasonableEffectivenessLarge2023} demonstrate the effectiveness of CLIP in the source-free video domain adaptation setting by directly adapting it to action recognition tasks despite fundamentally being an image-based technique. Instead, we leverage CLIP as a spatial teacher for training a spatiotemporal student model that can better operate on target domain videos.

This paper presents, to the best of our knowledge, the first exploration of self-supervised masked distillation techniques in the context of unsupervised video domain adaptation. Our approach, Unsupervised Adaptation with Teacher-Enhanced Learning (\methodname), successfully harnesses the capabilities of an image-based teacher in two ways: (1) to form the target representations in a masked self-supervised pre-training stage on target domain videos, and (2) to improve the quality of pseudolabels used during self-training in collaboration with the student video network (see \cref{fig:teaser}). \\

\noindent We summarize the contributions of our work as follows:

\begin{itemize}

\item We introduce the \methodname pipeline for video domain adaptation, which consists of a novel combination of masked video modeling and self-training techniques.

\item We conduct extensive experimentation by evaluating \methodname on three video domain adaptation benchmarks (\textit{Daily-DA}, \textit{Sports-DA} and \textit{UCF-HMDB}) and observe significant performance gains compared to previously reported results.

\item We present a series of ablation experiments that study the effectiveness of various aspects of the \methodname pipeline and assess alternative design choices. In particular, we show that masked distillation pre-training and masked self-training are best applied in concert to achieve the strongest domain transfer performance.

\end{itemize}

%% file: sec/2_related.tex
\section{Related Work}
\label{sec:related}

\textbf{Video Unsupervised Domain Adaptation (VUDA).} 
Techniques for video unsupervised domain adaptation typically fall within a few common categories \cite{xuVideoUnsupervisedDomain2022}. Adversarial methods attempt to align representations between the source and target domains by enforcing a domain confusion loss using a separate domain discriminator network, where the main network must learn to fool the discriminator. Some techniques which were developed for images, such as DANN \cite{ganin2016domain} or MK-MMD \cite{long2015learning}, can be applied directly to video architectures.  TA$^3$N \cite{chenTemporalAttentiveAlignment2019} is a video-specific approach that uses separate adversaries for the spatial and temporal dimensions. These approaches can result in unstable training due to competing objectives.

Another class of methods use contrastive learning and exploit the intrinsic structure of video. CoMix \cite{sahooContrastMixTemporal2021} uses contrastive learning with pseudolabeling and video-based augmentations such as sampling at varying frame rates or mixing the backgrounds of source videos and target videos. CO$^2$A \cite{turrisidacostaDualHeadContrastiveDomain2022} optimizes 6 different losses simultaneously, including contrastive losses at the clip- and video-level, along with supervised contrastive learning across domains. UDAVT \cite{dacostaUnsupervisedDomainAdaptation2022} adapts vision transformers using the information bottleneck principle by having separate projectors for source and target features and estimating a cross correlation matrix between features whose labels and pseudolabels match.

Some video domain adaptation techniques have been developed to address the related problem of source-free video video domain adaptation (SFVUDA), which focuses on scenarios where source data is unavailable during the adaptation phase. ATCoN \cite{xuSourceFreeVideoDomain2022} tries to ensure the trained model is as consistent with its features as the original model.  EXTERN \cite{xu2022extern}  additionally uses masking augmentations as a factor for features to be consistent over. DALL-V \cite{zaraUnreasonableEffectivenessLarge2023} takes a different approach by adapting a CLIP \cite{radfordLearningTransferableVisual2021a} image encoder to source and target datasets using an adapter.

\textbf{Masked Image \& Video Modeling.} 
Masked modeling first achieved success in natural language processing \cite{devlinBERTPretrainingDeep2019}, and has since been applied to image and video data as a way to learn powerful feature encoding from unlabeled data. The basic idea behind masked modeling is to partially occlude patches of the input data and train a model to reconstruct the missing patches. This can be done at the pixel-level \cite{heMaskedAutoencodersAre2021, liMSTMaskedSelfSupervised2021, xieSimMIMSimpleFramework2022, tongVideoMAEMaskedAutoencoders2022, wangVideoMAEV2Scaling2023, feichtenhoferMaskedAutoencodersSpatiotemporal2022}, or the targets of the reconstruction can be formed by a neural network--- either an exponential moving average (EMA) of the network being trained \cite{baevskiData2vecGeneralFramework2022, zhouImageBERTPretraining2022, assranMaskedSiameseNetworks2022, assranSelfSupervisedLearningImages2023, kakogeorgiouWhatHideYour2022, lin2023supervised}, or even a separate teacher model. MVD \cite{wangMaskedVideoDistillation2023} uses the latter concept to train a video transformer network using both image and video teachers, which can be viewed as a form of knowledge distillation \cite{hintonDistillingKnowledgeNeural2014}.

Unmasked Teacher (UMT) \cite{liUnmaskedTeacherTrainingEfficient2023} uses a different form of masked distillation to train powerful video encoders, where instead of reconstructing representations of masked regions, the student attempts to match the teacher representations of visible patches \cite{xueStareWhatYou2023}. UMT is able to use a spatial encoder (CLIP \cite{radfordLearningTransferableVisual2021a}) to train a powerful spatiotemporal model. Because the video encoder only needs to process visible patches, and there is no need for an expensive decoder for reconstruction, UMT offers a highly efficient approach for self-supervision on videos. As such, we choose to leverage the UMT masked distillation objective to perform unsupervised video learning in \methodname.

\textbf{Masked Modeling for Domain Adaptation.}
Masked modeling is fundamentally about ensuring a given feature is invariant to whether its corresponding input is masked or not. Inducing such invariances on novel domains can be a useful property to enforce for the purposes of domain adaptation. MAE-TTT \cite{gandelsman2022testtime} finetunes a masked autoencoder \cite{heMaskedAutoencodersAre2021} on single images at test time, adapting the feature extractor to the statistics of the current image while leaving the prediction head unchanged. MIC \cite{hoyerMICMaskedImage2023} uses an EMA teacher model on unmasked images to produce pseudolabels for training on masked target images. PACMAC \cite{prabhu2022adapting} selects target domain samples for pseudolabeling based on prediction consistency over masked views, and also applies the classification loss to masked target inputs.  Since these approaches have been developed for image-based tasks, there is still an open question of how similar concepts could be applied to video recognition.

%% file: sec/3_preliminaries.tex
\section{Preliminaries}
\label{sec:preliminaries}

\newcommand{\source}{S}
\newcommand{\target}{T}
\newcommand{\student}{a}
\newcommand{\teacher}{*}

\subsection{Problem Formulation}

We formalize the problem of unsupervised domain adaptation for action recognition as follows. Given a set of labeled videos $\mathcal{D}_\source := \{(\mathbf{x}^\source_i, y^\source_i)\}_{i=1}^{N_\source}$ drawn from a source data distribution $\mathcal{P}_\source$, and a set of unlabeled videos  $\mathcal{D}_\target :=\{\mathbf{x}^\target_i\}_{i=1}^{N_\target}$ drawn from a target distribution $\mathcal{P}_\target$, our objective is to learn a function that can successfully classify videos drawn from $\mathcal{P}_\target$. More precisely, we seek to learn the parameters $\theta$ of a function $f_{\theta}$, in our case a transformer neural network, that minimize the empirical risk on a sampling of videos from $\mathcal{P}_\target$. The challenge in domain adaptation lies in the fact that $\mathcal{P}_\source$ and $\mathcal{P}_\target$ differ from one another, leading to a gap in performance when directly applying a source trained model on target domain data. Thus, UDA techniques need to exploit information in unlabeled target domain samples to reduce this gap.

\subsection{Self-Supervised Initialization}
\label{sec:initialization}

In contrast to some other video domain adaptation works \cite{dacostaUnsupervisedDomainAdaptation2022, xuAugmentingAligningSnippets2023} that use network weights from Kinetics-400 supervised pre-training, we initialize our network from \textit{self-supervised} pre-training. As contended in \cite{prabhu2022adapting} for UDA with images, we argue that supervised pre-training can potentially complicate the study of VUDA techniques. When the pre-trained network has been trained to classify categories present in the DA dataset, some DA techniques could perform well simply by preserving the capabilities of the pre-trained model despite not generalizing well to DA tasks that do not have category overlap with the pre-training dataset.
Moreover, the UDA problem formulation implies that labeled instances of relevant classes are only available from the source domain. This condition does not hold for the \textit{Daily-DA} and \textit{Sports-DA} benchmarks that we use in our evaluations, as 6 out of their 8 classes also appear in the Kinetics-400 dataset. Further exacerbating the issue with supervised Kinetics pre-training is the fact that Kinetics serves as a target domain in both benchmarks. Thus, we initialize the network weights from UMT \cite{liUnmaskedTeacherTrainingEfficient2023} single-modality (\ie, video only) self-supervised pre-training on Kinetics-710 \cite{liUniFormerV2SpatiotemporalLearning2023}. Please refer to \cref{sec:k400_init_results_supp} in the Appendix for select results using a UMT pre-trained network with additional supervised fine-tuning on Kinetics-400, which unsurprisingly achieves better baseline performance.

%% file: sec/4_method.tex
\section{Method}
\label{sec:method}

\newenvironment{zeroindent}
  {\par\setlength{\parindent}{0pt}}
  {\par}

\begin{figure}[t]
  \centering
    \includegraphics[width=0.95\linewidth]{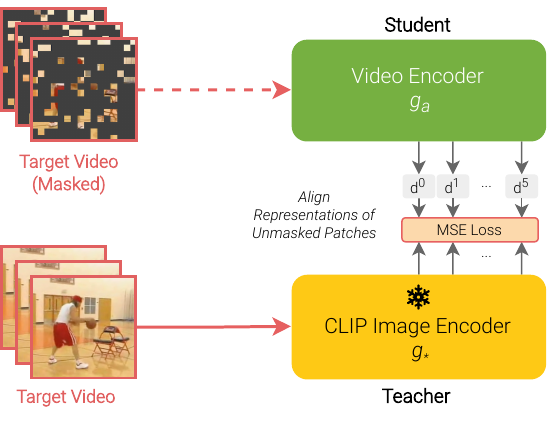}

   \caption{Unmasked Teacher (UMT) training used in Stage 1 of \methodname to perform unsupervised representation learning on target domain videos.}
   \label{fig:umt}
\end{figure}

\begin{figure*}[t]
  \centering
    \includegraphics[width=0.7\linewidth]{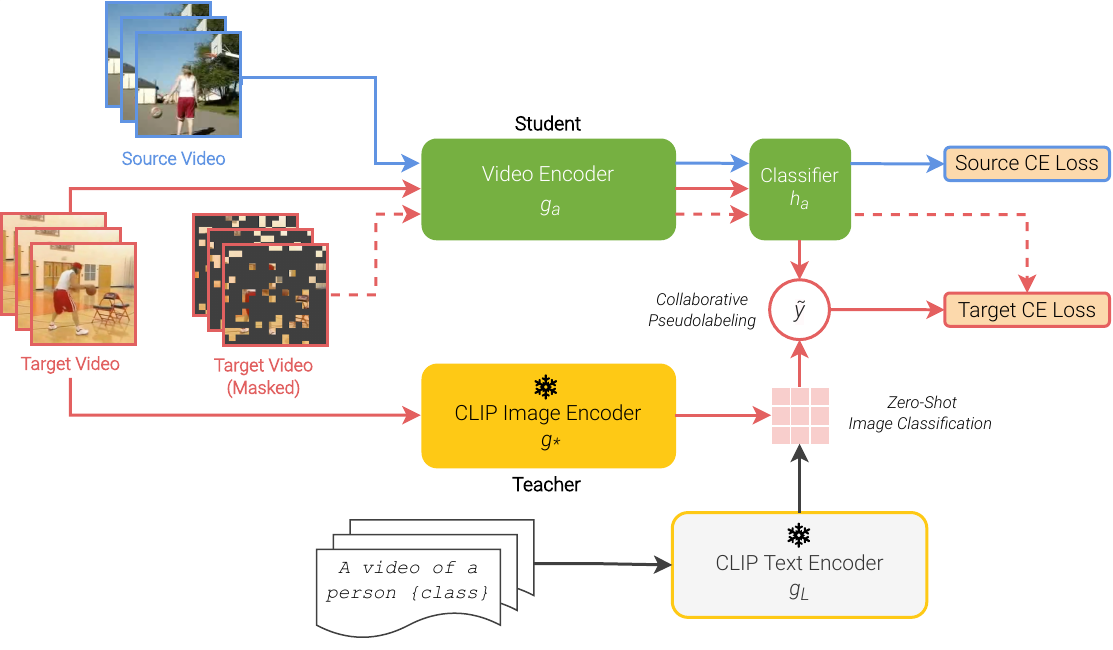}

   \caption{Overview of the collaborative self-training stage in \methodname. The student and teacher models work together to produce more accurate pseudolabels for target domain videos. The target domain classification loss is enforced on masked target videos to encourage stronger context learning. Source domain classification is included to stabilize training, which is especially important at the beginning of training when pseudolabels may have low accuracy.}
   \label{fig:method}

\end{figure*}

\begin{zeroindent}
\subsection{\methodname}

Our approach to VUDA consists of three stages: (1) unsupervised pre-training on target domain data, (2) supervised fine-tuning on source domain data and (3) collaborative self-training using videos from both domains. We now describe each stage in detail. \\

\textbf{Stage 1: Unsupervised Target Domain Pre-Training}.

To enable the extraction of discriminative features from target domain videos, the \methodname pipeline begins by performing additional unsupervised pre-training on the target dataset $\mathcal{D}_\target$. For this, we adopt the UMT \cite{liUnmaskedTeacherTrainingEfficient2023} self-supervised training objective. UMT training uses a pre-trained and frozen spatial teacher model $g_{\teacher}$ to train a spatiotemporal student model $g_{\student}$ by enforcing alignment between feature representations of the two networks at multiple layers (depicted in \cref{fig:umt}). A key aspect of the training process is that the student model takes as input a masked video view $m(\mathbf{x}^\target)$, while the teacher model processes unmasked video frames from $\mathbf{x}^\target$. Through UMT training, $g_{\student}$ must learn to match the semantically rich (albeit only spatially informed) features of $g_{\teacher}$ from heavily masked target domain videos. To ensure sampling of meaningful video patches, UMT performs an attention-guided masking operation $m(\cdot)$ drawn from a multinomial distribution defined by the attention map of the final layer CLS token in $g_{\teacher}$. Attention-guided masking allows the use of a high masking ratio, which makes UMT training extremely efficient.

We formalize the Stage 1 objective as follows. Let \( \mathbf{z}^l_\student \) denote the L2 normalized \( l \)-th layer representation of $g_{\student}$ for a masked input \( m(\mathbf{x}) \), and let \( \mathbf{z}^l_\teacher \) denote the normalized \( l \)-th layer representation of  $g_{\teacher}$ for the full input \( \mathbf{x} \), but only at the locations corresponding to the visible patches in \( m(\mathbf{x}) \).
The loss function in Stage 1 can be represented as:
\[
\mathcal{L}_{\text{UMT}} = \mathbb{E}_{\mathbf{x} \sim \mathcal{P}_\target} \left[ \frac{1}{|\mathcal{A}|} \sum_{l \in \mathcal{A}} \text{MSE}\left( d^l(\mathbf{z}^l_\student), \mathbf{z}^l_\teacher \right) \right]
\]

where \( \mathcal{A} \) is a set of aligned layers, $d^l(\cdot)$ is a linear projection for the \( l \)-th layer and MSE indicates mean squared error.

Following the original UMT method, we leverage a CLIP \cite{radfordLearningTransferableVisual2021a} ViT-B image encoder as $g_{\teacher}$. We keep the standard UMT masking ratio of $r=0.8$ and perform alignment between the last 6 layers of the networks. $g_{\student}$ is initialized as described in \cref{sec:initialization}, along with the projections $d^l(\cdot)$, which are kept frozen throughout Stage 1. \\

\textbf{Stage 2: Source Domain Fine-Tuning}.
The second stage of \methodname fits a classifier on source domain data using $g_{\student}$ from Stage 1. We introduce label space information in a separate stage from target domain self-supervision because, as the authors of \cite{gandelsman2022testtime} observe for MAE training, we find that imposing a classification loss and UMT loss simultaneously can hinder convergence. To perform supervised fine-tuning, we introduce a linear classification head $h_{\student}$ that operates on the mean pooled outputs of the final transformer layer in $g_{\student}$. The full student classifier network can now be represented as $f_{\student} = h_{\student}(g_{\student}(\cdot))$. As is the case with other masked pre-training strategies, we find that representations learned via UMT pre-training are not linearly separable. Thus, we perform full fine-tuning of $f_{\student}$ by minimizing a cross entropy (CE) loss $\mathcal{L}_{CE}$ on unmasked source domain videos. The Stage 2 loss function is represented as:

\begin{equation}
\mathcal{L}_{\text{SFT}} = \mathbb{E}_{(\mathbf{x}^\source, y^\source) \sim \mathcal{P}_\source} \left[ \mathcal{L}_{CE}(f_{\student}(\mathbf{x}^\source), y^\source) \right]
\end{equation}

To help preserve target domain feature extraction learned in Stage 1, we apply a layer-wise learning rate decay when fine-tuning on source domain data. \\

\noindent\textbf{Stage 3: Collaborative Self-Training (CST)}.
After enhancing the ability of $g_\student$ to extract meaningful features in the target domain (Stage 1) and fitting $f_\student$ for the classification task using source videos (Stage 2), we further adapt $f_\student$ to target data using a self-training process (depicted in \cref{fig:method}). In a conventional self-training framework (denoted as \( \mathcal{L}_\text{ST} \) in \cref{eqn:conventional_st}), the overall loss is a sum of losses on labeled source domain samples and select target domain samples, chosen using mask \( s(\cdot) \). The target domain loss leverages the model's own predictions $\hat{y}$ (where $\hat{y} = {\mathrm{argmax}}_c\, \sigma(f(\mathbf{x}^\target))[c]$ and $\sigma(\cdot)$ denotes the softmax function) in lieu of labels in the CE loss computation. Various strategies have been used to form the selection mask $s(\cdot)$, with the goal of choosing target samples whose predictions are more likely to be correct \cite{sohnFixMatchSimplifyingSemiSupervised2020, zhang2021flexmatch, prabhu2022adapting}.

\begin{equation}
\begin{split}
\mathcal{L}_{\text{ST}} = & \mathbb{E}_{(\mathbf{x}^\source, y^\source) \sim \mathcal{P}_\source} \left[ \mathcal{L}_{CE}(f_{\student}(\mathbf{x}^\source), y^\source) \right] + \\
\lambda & \mathbb{E}_{\mathbf{x}^\target \sim \mathcal{P}_\target} \left[ s(\mathbf{x}^\target)\mathcal{L}_{CE}(f_{\student}(\mathbf{x}^\target), \hat{y}_\student(\mathbf{x}^\target)) \right]
\end{split}
\label{eqn:conventional_st}
\end{equation}

We introduce a modified form of self-training, which we refer to as collaborative self-training (CST), that makes use of the student-teacher setup from Stage 1 to improve the process of pseudolabel estimation. Specifically, we use the zero-shot classification capabilities of the CLIP image teacher model $f_\teacher$ (see \cref{sec:imp_deets} for details), in combination with predictions from the video student model $f_\student$, to create refined target pseudolabels for self-training. We employ the MatchOrConf scheme proposed in \cite{zhangRethinkingRolePreTrained2023} to combine the outputs of the two models as shown in \cref{eqn:matchORconf}, where $\text{Conf}(\hat{y})$ denotes the maximum softmax probability \cite{hendrycks2016baseline}, $\max_c \sigma(f(\mathbf{x}^\target))[c]$, to create pseudolabel $\tilde{y}$:

\begin{equation}
\tilde{y} =
\begin{cases}
\hat{y}_\student & \text{if } \hat{y}_\student = \hat{y}_\teacher \\
\hat{y}_\student & \text{if } \hat{y}_\student \neq \hat{y}_\teacher \text{ and } \text{Conf}(\hat{y}_\student) > \gamma \text{ and } \text{Conf}(\hat{y}_\teacher) \leq \gamma, \\
\hat{y}_\teacher & \text{if } \hat{y}_\student \neq \hat{y}_\teacher \text{ and } \text{Conf}(\hat{y}_\student) \leq \gamma \text{ and } \text{Conf}(\hat{y}_\teacher) > \gamma, \\
-1 & \text{otherwise}.
\end{cases}
\label{eqn:matchORconf}
\end{equation}

$\mathbf{x}^T$ is an argument of the above, but is omitted for clarity. The MatchOrConf scheme selects only those target samples where there is agreement between the predictions of the two models, or where one model's confidence exceeds threshold $\gamma$ while the other's does not, resulting in the following selection mask:

\begin{equation}
s(\mathbf{x}^\target) = 
\begin{cases}
1 & \text{if } \tilde{y}(\mathbf{x}^\target) \neq -1, \\
0 & \text{otherwise}.
\end{cases}
\end{equation}

Taking inspiration from \cite{hoyerMICMaskedImage2023} and \cite{prabhu2022adapting}, we compute the target domain loss on masked videos, where we employ the same teacher-guided attention masking strategy (with $r=0.8$) used during UMT pre-training in Stage 1. We find that training on masked videos in this stage results in substantial performance boosts compared to training on unmasked videos (see \cref{tab:pl_target_masking}). Like \cite{hoyerMICMaskedImage2023}, we include a loss weighting $q$ for each target video, based on the student network confidence:

\begin{equation}
q(\mathbf{x}^\target) = \text{Conf}(\hat{y}_\student(\mathbf{x}^\target))
\end{equation}

The complete loss function for the collaborative self-training stage can be expressed as:

\begin{equation}
\begin{split}
\mathcal{L}_{\text{CST}} = & \mathbb{E}_{(\mathbf{x}^\source, y^\source) \sim \mathcal{P}_\source} \left[ \mathcal{L}_{CE}(f_{\student}(\mathbf{x}^\source), y^\source) \right] + \\ 
\lambda & \mathbb{E}_{(\mathbf{x}^\target) \sim \mathcal{P}_\target} \left[ s(\mathbf{x}^\target)q(\mathbf{x}^\target)\mathcal{L}_{CE}(f_{\student}(m(\mathbf{x}^\target)), \tilde{y}(\mathbf{x}^\target) \right]
\end{split}
\end{equation}

\end{zeroindent}

%% file: sec/5_experiments.tex
\section{Experiments}
\label{sec:experiments}

%%%%%%%%%%%%%%%%%%%%%% DATASETS %%%%%%%%%%%%%%%%%%%%%%

\subsection{Datasets}

We evaluate our approach on three video domain adaptation benchmarks: {\textit{Daily-DA}}, {\textit{Sports-DA}}, {\textit{UCF$\leftrightarrow$HMDB}\(_{full}\)}.

\begin{itemize}
\item \noindent\textbf{{\textit{Daily-DA}}} \cite{xuMultiSourceVideoDomain2023} is a recently introduced benchmark composed of videos from four domains: ARID (A) \cite{xuARIDNewDataset2022}, HMDB51 (H) \cite{kuehneHMDBLargeVideo2011}, Moments-in-Time (M) \cite{monfortMomentsTimeDataset2020} and Kinetics (K) \cite{kayKineticsHumanAction2017, carreiraShortNoteKinetics6002018}. It consists of 8 overlapping classes representing everyday actions, with a total volume of approximately 19K videos. Videos from the ARID domain present a unique challenge in this benchmark, as they were explicitly filmed under low-illumination conditions.

\item \noindent\textbf{{\textit{Sports-DA}}} \cite{xuMultiSourceVideoDomain2023} contains videos from three domains: Sports-1M (S) \cite{karpathyLargeScaleVideoClassification2014}, UCF101 (U) \cite{soomroUCF101Dataset1012012} and Kinetics (K) \cite{kayKineticsHumanAction2017, carreiraShortNoteKinetics6002018}. It includes 23 overlapping categories depicting various sporting activities, with a total volume of approximately 41K videos.

\item \textbf{{\textit{UCF$\leftrightarrow$HMDB}\(_{full}\)}} \cite{chenTemporalAttentiveAlignment2019} contains approximately 3.2K videos across 12 action categories taken from the UCF101 (U) \cite{soomroUCF101Dataset1012012} and HMDB51 (K) \cite{kuehneHMDBLargeVideo2011} datasets.
\end{itemize}

\input{tables/uda_dailyDA}
\input{tables/uda_sportsDA}

\subsection{Implementation Details}
\label{sec:imp_deets}

\noindent\textbf{Frame Sampling.}
During training, we perform random uniform frame sampling \cite{wangTemporalSegmentNetworks2017} on each video. We divide the video into $T=8$ uniform segments and randomly sample one frame from each. All frames are resized to 224 $\times$ 224 pixels, resulting in a 3 $\times$ 8 $\times$ 224 $\times$ 224 tensor for each video. During testing, we average the network predictions across 3 spatial crops and 4 temporal clips, where clip $i$ is formed by splitting each of the $T$ segments into 4 sub-segments and sampling a frame from the $i$-th sub-segment of each.

\noindent\textbf{Network Architecture \& Training.}
All of our experiments use a standard ViT-B/16 \cite{dosovitskiyImageWorth16x162021} architecture with a fixed sine and cosine positional encoding scheme. Our patch embedding projection operates on each frame independently, and self-attention is performed across both the space and time dimensions. All classification is performed using mean pooling of the final layer patch representations. We conduct our experiments using the PyTorch deep learning library \cite{paszke2019pytorch} on 4 NVIDIA A5000 GPUs.

\noindent During Stage 1, we perform UMT training for 50 epochs on the target dataset using a batch size of 256 and a learning rate of 1.5e-5. 
During Stage 2, we use a batch size of 28, a learning rate of 2.5e-5, and apply a layer-wise learning rate decay of 0.65. 
During stage 3, we use a batch size of 40 (20 videos from each domain), a learning rate of 1e-5, a MatchOrConf confidence threshold of $\gamma=0.1$ and a target domain loss weight $\lambda=1$. 
Training in all stages uses the AdamW optimizer \cite{loshchilov2017decoupled} and a weight decay of 0.05. Please refer to \cref{sec:imp_deets_supp} in the Appendix for more training details for each stage.

\noindent\textbf{Zero-Shot Classification with CLIP}. In Stage 3 of \methodname, we use CLIP (ViT-B/16) in a zero-shot manner to refine the pseudolabels of target domain videos. We do so by passing each of the $T$ video frames through the CLIP image encoder and computing the cosine similarity with a set of text representations produced by the CLIP text encoder. The inputs to the text encoder are the class names associated with the particular action recognition task, embedded in the following template: ``\textit{A video of a person} \{class\}." To form a video-level prediction, we simply average the softmax of the cosine similarity scores across the video frames. Refer to the Appendix (\cref{sec:clip_supp}) for a listing of class names used for each task.

\subsection{Baselines}

We compare results using \methodname with previously developed techniques in UDA, as well as in the closely related setting of SFUDA as reported by \cite{zaraUnreasonableEffectivenessLarge2023}. For \textit{Daily-DA} and \textit{Sports-DA} we compare against DANN \cite{ganin2016domain}, MK-MMD \cite{long2015learning} and TA$^3$N \cite{chenTemporalAttentiveAlignment2019} in the UDA setting, and ATCoN \cite{xuSourceFreeVideoDomain2022}, EXTERN \cite{xu2022extern} and DALL-V \cite{zaraUnreasonableEffectivenessLarge2023} for SFUDA. For \textit{UCF$\leftrightarrow$HMDB}\(_{full}\), we additionally report accuracies from CO$^2$A \cite{turrisidacostaDualHeadContrastiveDomain2022} and UDAVT \cite{dacostaUnsupervisedDomainAdaptation2022}. 

\subsection{Benchmark Results}

Tables \ref{tab:uda_dailyDA}, \ref{tab:uda_sportsDA} and \ref{tab:uda_ucf-hmdb} show the results of our evaluation of \methodname on \textit{Daily-DA}, \textit{Sports-DA} and \textit{UCF$\leftrightarrow$HMDB}\(_{full}\). 
We include accuracies after Stage 2 of the pipeline (denoted as \methodname w/o CST) as well as after applying the full \methodname process.
For reference, we include accuracies using our UMT pre-trained network with source domain supervised fine-tuning (Source Only) and with target domain supervised fine-tuning (Target Only). We also include the source only and target only accuracies (denoted in the tables as LB and UB) of DALL-V \cite{zaraUnreasonableEffectivenessLarge2023} (a CLIP-based ResNet-50 model) for added context.
We additionally display the zero-shot classification accuracy of the CLIP teacher model that we utilize during Stage 1 and Stage 3 of \methodname.

% Daily-DA
An observation we make across all benchmarks is the enhanced baseline accuracy of the UMT pre-trained model, which speaks to the strength of the unsupervised masked distillation process and the ViT architecture. On the \textit{Daily-DA} benchmark, we find that \methodname exceeds previously reported results on most domain shifts. On only one shift (M$\rightarrow$A) we observe degradation in target accuracy after employing \methodname compared to source only, which is consistent with previous work.

% Sports-DA
\methodname exhibits even stronger performance on the larger-scale \textit{Sports-DA} benchmark, where it exceeds previous results by a large margin. We also see more consistent and substantial improvement after the UMT target pre-training stage alone (\methodname w/o CST), suggesting that overtraining may be occurring on the smaller datasets in \textit{Daily-DA}.

%HMDB-UCF
On {\textit{UCF$\leftrightarrow$HMDB}\(_{full}\)} we see state-of-the-art performance on U$\rightarrow$H, while UDAVT outperforms on H$\rightarrow$U. Nevertheless, \methodname still achieves strong overall performance on the benchmark.

\input{tables/uda_ucf-hmdb}

%% file: tables/uda_dailyDA.tex
\begin{table*}[!ht]
\begin{center}
\small
\resizebox{1.0\linewidth}{!}{
\begin{tabular}{clccc|ccc|ccc|ccc|c}
\toprule
& \multirow{2}{*}{\textbf{Method}} & \multicolumn{12}{c}{\textbf{Target Domain Accuracy (Top-1\%)}} \\
& & H→A & M→A & K→A & A→H & M→H & K→H & H→M & A→M & K→M & M→K & H→K & A→K &  \textbf{Avg.} \\
\midrule
& DALL-V \cite{zaraUnreasonableEffectivenessLarge2023} LB & 17.5 & 34.7 & 15.6 & 14.6 & 44.6 & 47.9 & 25.5 & 15.5 & 35.7 & 61.6 & 45.1 & 17.8 & 31.3 \\
& \socolor Source Only & \socolor 40.4 & \socolor \textbf{52.1} & \socolor 36.5 & \socolor 49.6 & \socolor 68.3 & \socolor 57.9 & \socolor 41.5 & \socolor 36.3 & \socolor 43.3 & \socolor 79.3 & \socolor 48.0 & \socolor 41.7 & \socolor 49.6 \\
\midrule    
\scriptsize \parbox[t]{2mm}{\multirow{1}{*}{\rotatebox{90}{ZS}}}
& CLIP (ViT-B/16) \cite{radfordLearningTransferableVisual2021a} & 36.5 & 36.5 & 36.5 & 60.0 & 60.0 & 60.0 & 48.5 & 48.5 & \textbf{48.5} & 68.1 & 68.1 & 68.1 & 53.3 \\
\midrule
\scriptsize \parbox[t]{2mm}{\multirow{3}{*}{\rotatebox{90}{SFVUDA}}}
& ATCoN \cite{xuSourceFreeVideoDomain2022} & 17.9 & 27.2 & 17.2 & 26.7 & 47.3 & 48.2 & 30.7 & 17.2 & 32.5 & 57.7 & 48.5 & 31.0 & 33.5 \\
& EXTERN \cite{xu2022extern} & 26.2 & 18.1 & 23.9 & 26.2 & 53.7 & 55.8 & 40.7 & 18.2 & 35.2 & 68.1 & 57.6 & 51.4 & 39.6 \\
& DALL-V \cite{zaraUnreasonableEffectivenessLarge2023} & 24.0 & 24.0 & 24.0 & 57.9 & 65.4 & 52.5 & 47.0 & 45.7 & 47.0 & 78.1 & \textbf{76.7} & \textbf{75.0} & 51.4 \\
\midrule
\scriptsize \parbox[t]{2mm}{\multirow{5}{*}{\rotatebox{90}{UDA}}} & DANN \cite{ganin2016domain} & 14.2 & 22.8 & 21.2 & 20.1 & 43.3 & 37.5 & 29.5 & 19.7 & 21.7 & 58.8 & 38.2 & 27.0 & 29.5 \\
& MK-MMD \cite{long2015learning} & 20.3 & 21.0 & 21.7 & 18.7 & 50.4 & 36.2 & 25.7 & 18.0 & 24.0 & 58.5 & 33.8 & 26.1 & 29.5 \\
& TA$^3$N \cite{chenTemporalAttentiveAlignment2019} & 14.4 & 21.6 & 19.9 & 14.9 & 43.0 & 37.7 & 25.7 & 15.6 & 31.5 & 55.5 & 38.4 & 23.4 & 28.5 \\
& \ourcolor \methodname w/o CST & \ourcolor 43.8 & \ourcolor {46.7} & \ourcolor 34.7 & \ourcolor 51.7 & \ourcolor 70.8 & \ourcolor 54.6 & \ourcolor 44.3 & \ourcolor 39.0 & \ourcolor 43.3 & \ourcolor 78.1 & \ourcolor 51.7 & \ourcolor 50.8 & \ourcolor 50.8 \\
& \ourcolor \methodname (Ours) & \ourcolor \textbf{48.0} & \ourcolor 44.1 & \ourcolor \textbf{37.5} & \ourcolor \textbf{67.9} & \ourcolor \textbf{74.2} & \ourcolor \textbf{65.8} & \ourcolor \textbf{51.8} & \ourcolor \textbf{50.0} & \ourcolor 48.0 & \ourcolor \textbf{89.9} & \ourcolor 69.9 & \ourcolor 63.6 & \ourcolor \textbf{59.2} \\
\midrule
& DALL-V  \cite{zaraUnreasonableEffectivenessLarge2023} UB & 26.9 & 26.9 & 26.9 & 70.4 & 70.4 & 70.4 & 61.5 & 61.5 & 61.5 & 88.9 & 88.9 & 88.9 & 61.9 \\
& \tocolor Target Only & \tocolor 68.5 & \tocolor 68.5 & \tocolor 68.5 & \tocolor 84.6 & \tocolor 84.6 & \tocolor 84.6 & \tocolor 73.0 & \tocolor 73.0 & \tocolor 73.0 & \tocolor 98.3 & \tocolor 98.3 & \tocolor 98.3 & \tocolor 81.1 \\
\bottomrule
\end{tabular}}
\end{center}
\vspace{-4mm}
\caption{UDA results on \textit{Daily-DA}. Rows in color use our UMT pre-trained backbone, and rows without color are reported from \cite{xuAligningCorrelationInformation2022}.}
\label{tab:uda_dailyDA}
\end{table*}

%% file: tables/uda_sportsDA.tex
\begin{table*}[ht]
\small
\centering
\resizebox{0.6\linewidth}{!}{
\begin{tabular}{clcc|cc|cc|c}
\toprule
& \multirow{2}{*}{\textbf{Method}} & \multicolumn{5}{c}{\textbf{Target Domain Accuracy (Top-1 \%)}} \\
& & U→S & K→S & S→U & K→U & U→K & S→K & \textbf{Avg.} \\
\midrule
& DALL-V  \cite{zaraUnreasonableEffectivenessLarge2023} LB & 64.3 & 79.5 & 84.4 & 85.4 & 67.2 & 78.2 & 76.5 \\ 
& \socolor Source Only & \socolor 67.6 & \socolor 86.8 & \socolor 97.9 & \socolor 98.9 & \socolor 79.9 & \socolor 89.1 & \socolor 86.7 \\ 

\midrule
\scriptsize \parbox[t]{2mm}{\multirow{1}{*}{\rotatebox{90}{ZS}}}
& CLIP (ViT-B/16)~\cite{radfordLearningTransferableVisual2021a} & 85.0 & 85.0 & 93.3 & 93.3 & 91.3 & 91.3 & 89.9 \\
\midrule

\scriptsize \parbox[t]{2mm}{\multirow{3}{*}{\rotatebox{90}{SFVUDA}}} & ATCoN~\cite{xuSourceFreeVideoDomain2022} & 47.9 & 69.7 & 90.6 & 93.6 & 65.2 & 76.0 & 73.8 \\
& EXTERN~\cite{xu2022extern} & 72.7 & 73.8 & 95.4 & 93.7 & 81.2 & 82.2 & 83.2 \\
& DALL-V~\cite{zaraUnreasonableEffectivenessLarge2023} & 75.9 & 77.7 & 88.8 & 88.0 & 81.2 & 82.3 & 82.3 \\
\midrule
\scriptsize \parbox[t]{2mm}{\multirow{5}{*}{\rotatebox{90}{UDA}}} & DANN~\cite{ganin2016domain} & 55.1 & 75.0 & 85.7 & 88.0 & 65.9 & 73.4 & 73.8 \\
& MK-MMD~\cite{long2015learning} & 55.6 & 67.9 & 90.9 & 90.2 & 66.1 & 73.6 & 74.0 \\
& TA$^3$N~\cite{chenTemporalAttentiveAlignment2019} & 54.1 & 68.6 & 93.0 & 90.3 & 63.6 & 72.6 & 73.7 \\

& \ourcolor \methodname w/o CST & \ourcolor 73.9 & \ourcolor 87.3 & \ourcolor 98.2 & \ourcolor 99.2 & \ourcolor 84.6 & \ourcolor 90.2 & \ourcolor 88.9 \\
& \ourcolor \methodname~(Ours)& \ourcolor \textbf{86.0} & \ourcolor \textbf{90.2} & \ourcolor \textbf{98.7} & \ourcolor \textbf{99.8} & \ourcolor \textbf{94.2} & \ourcolor \textbf{95.3} & \ourcolor \textbf{94.0} \\

\midrule
& DALL-V  \cite{zaraUnreasonableEffectivenessLarge2023} UB & 88.3 & 88.3 & 93.4 & 93.4 & 85.6 & 85.6 & 89.1 \\ 
& \tocolor Target Only & \tocolor 97.9 & \tocolor 97.9 & \tocolor 98.8 & \tocolor 98.8 & \tocolor 99.9 & \tocolor 99.9 & \tocolor 98.9 \\ 

\bottomrule
\end{tabular}}
\caption{UDA results on {\textit{Sports-DA}}. Rows in color use our UMT pre-trained backbone, and rows without color are reported from \cite{xuAligningCorrelationInformation2022}.
}
\label{tab:uda_sportsDA}
\end{table*}

%% file: tables/uda_ucf-hmdb.tex
\begin{table}[t]
\begin{center}
\resizebox{0.72\linewidth}{!}{
\begin{tabular}{clcc|c}
\toprule
& \multirow{2}{*}{\textbf{Method}} & \multicolumn{3}{c}{\textbf{Accuracy (\%)}} \\
& & H→U & U→H & \textbf{Avg.} \\
\midrule
&  DALL-V  \cite{zaraUnreasonableEffectivenessLarge2023} LB & 71.6 & 76.1 & 73.8 \\
&  \socolor Source Only & \socolor 88.8 & \socolor 80.0 & \socolor 84.4 \\
\midrule
\scriptsize \parbox[t]{2mm}{\multirow{1}{*}{\rotatebox{90}{ZS}}}
& CLIP (ViT-B/16) \cite{radfordLearningTransferableVisual2021a} & 88.8 & 91.7 & 90.3 \\
\midrule
\scriptsize \parbox[t]{2mm}{\multirow{3}{*}{\rotatebox{90}{SFVUDA}}} & ATCoN \cite{xuSourceFreeVideoDomain2022} & 85.3 & 79.7 & 82.5 \\
& EXTERN \cite{xu2022extern} & 91.9 & {88.9} & 90.4 \\
& DALL-V \cite{zaraUnreasonableEffectivenessLarge2023} & 93.1 & {88.9} & {91.0} \\
\midrule
\scriptsize \parbox[t]{2mm}{\multirow{7}{*}{\rotatebox{90}{UDA}}} & DANN \cite{ganin2016domain} & 74.4 & 75.1 & 74.8 \\
& MK-MMD \cite{long2015learning} & 74.7 & 79.7 & 77.2 \\
& TA$^3$N \cite{chenTemporalAttentiveAlignment2019} & 78.1 & 84.8 & 81.5 \\
& CO$^2$A \cite{turrisidacostaDualHeadContrastiveDomain2022} & 95.8 & 87.8 & 91.8 \\
& UDAVT \cite{dacostaUnsupervisedDomainAdaptation2022} & \textbf{96.8} & 92.3 & \textbf{94.6} \\
& \ourcolor  \methodname w/o CST & \ourcolor 92.1 & \ourcolor 83.6 & \ourcolor 87.9 \\
& \ourcolor \methodname~(Ours) & \ourcolor 92.5 & \ourcolor \textbf{95.0} & \ourcolor 93.8 \\
\midrule
&  DALL-V  \cite{zaraUnreasonableEffectivenessLarge2023} UB & 93.7 & 91.4 & 92.6 \\
& \tocolor Target Only & \tocolor 99.7 & \tocolor 96.7 & \tocolor 98.2 \\
\bottomrule
\end{tabular}}
\end{center}
\vspace{-6mm}
\caption{UDA results on {\textit{UCF$\leftrightarrow$HMDB}\(_{full}\)}. Rows in color use our UMT backbone, and rows without color are reported from \cite{xuAligningCorrelationInformation2022}.}
\vspace{-2mm}
\label{tab:uda_ucf-hmdb}
\end{table}

%% file: sec/6_analysis.tex
\section{Additional Analysis \& Discussion}
\label{sec:analysis}

\begin{figure}[t]
  \centering
    \includegraphics[width=0.9\linewidth]{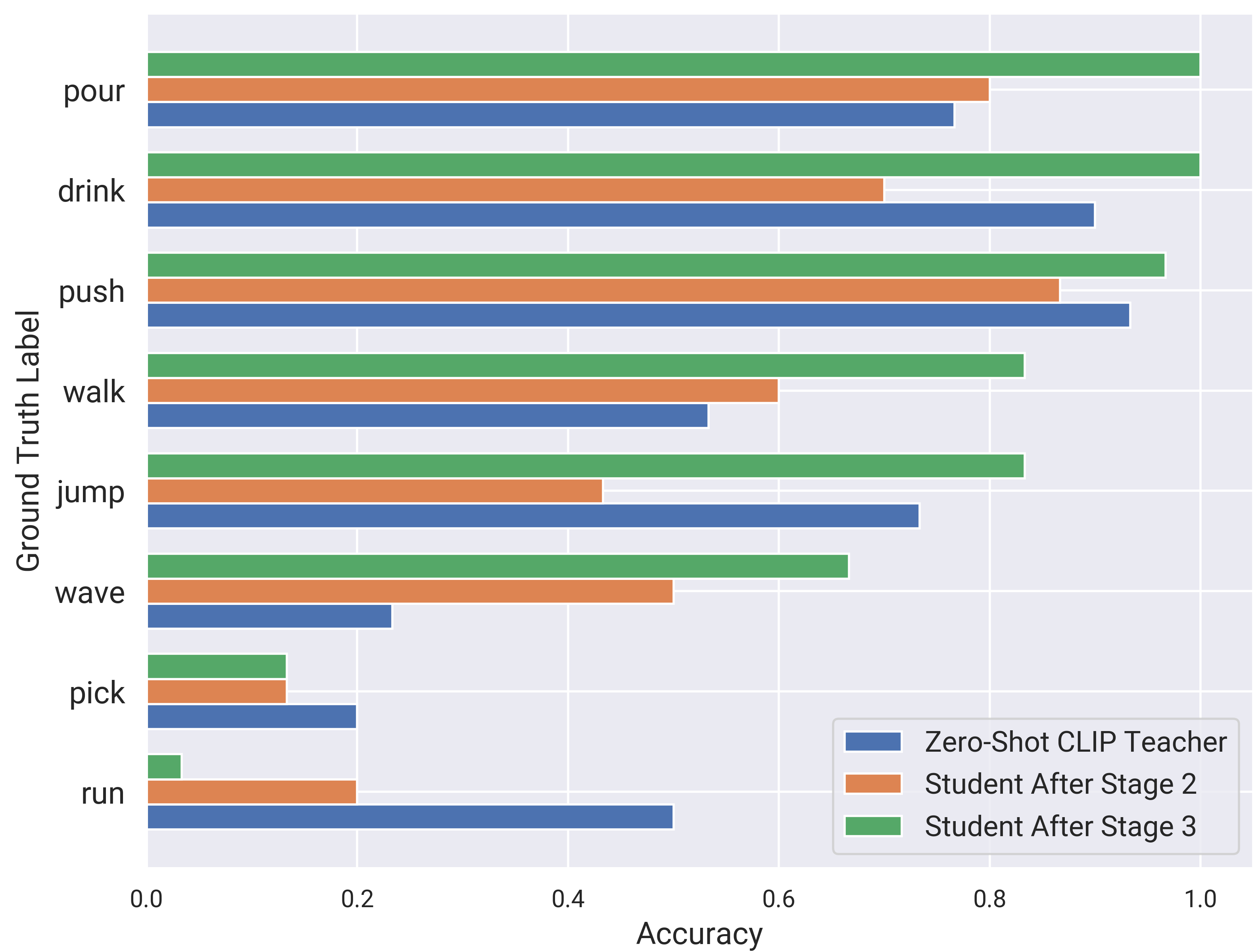}

   \caption{Class-wise accuracy on ARID$\rightarrow$HMDB. Performances are shown for the CLIP teacher model and the student model before and after the collaborative self-training stage (Stage 3). For most classes, the Stage 3 student exceeds the accuracy of both the Stage 2 student and the CLIP teacher.}
   \label{fig:collab_classwise_acc}
\end{figure}

\noindent\textbf{Student exceeds the teacher.}
We see in the benchmark results that the video student model exceeds the zero-shot accuracy of the CLIP teacher in virtually all domain shifts. This suggests that \methodname successfully leverages the spatial modeling capabilities of CLIP to train an even more powerful spatiotemporal model for the action recognition task. We can see from \cref{fig:collab_classwise_acc} that the teacher model and the student model prior to collaborative self-training have differing capabilities. For example, the image teacher is stronger at recognizing actions from the \textit{jump} class than the Stage 2 student. On the other hand, the video student is significantly better than the teacher at recognizing \textit{wave} at the start of CST. In both cases, we see that the student model, through collaborative self-training, comes to exceeds both the Stage 2 student and the CLIP image teacher. With the MatchOrConf pseudolabeling scheme, the two models are able to work together to achieve higher accuracy in the target domain. We see accuracy on the \textit{pick} and \textit{run} classes declines after CST, as self-training tends to reinforce the accuracy of well-performing classes, sometimes at the expense of more difficult ones that are infrequently assigned pseudolabels. Modifying CST to account for the learning status of each class, like in \cite{zhang2021flexmatch}, may help remedy this issue.

% STAGE ABLATION
\noindent\textbf{Impact of Stage 1 and 3}. 
In \cref{tab:stage_ablation} we ablate the stages in \methodname. 
While UMT pre-training and collaborative self-training each offer a benefit in the absence of the other, the largest gains in domain transfer performance are obtained when combining the two stages. 
For example, on the ARID$\rightarrow$HMDB domain shift in \textit{Daily-DA}, Stage 1 and Stage 3 alone improve upon source only accuracy by +1.8\% and +11.2\% respectively. 
Meanwhile, the full \methodname pipeline results in +18.3\% accuracy improvement on the target domain, suggesting that the masked pre-training and masked self-training processes complement one another.

\input{tables/ab_method}

\noindent\textbf{Impact of data domains during pre-training.} 
In \cref{tab:umt_data_ablation}, we assess the impact of utilizing each of the two data domains during UMT pre-training on UDA performance. 
While the results vary across domain shifts, we observe the most consistent improvements when pre-training only on target domain videos.
Although incorporating both domains during unsupervised pre-training has been successfully applied in image-based domain adaptation \cite{shenConnectNotCollapse2022, prabhu2022adapting, kim2021cds, mishraSurprisinglySimpleSemiSupervised2021}, we find it to be suboptimal with UMT training on the VUDA tasks studied in this work.

\input{tables/ab_pretrain}

% Masked Target During Self-Training
\noindent\textbf{Effect of masking during self-training.}
In \cref{tab:pl_target_masking}, we see that enforcing the target domain CE loss on masked videos results in increased performance compared to using unmasked target videos, with smaller benefits for H$\rightarrow$A and much larger benefits for A$\rightarrow$H. We believe the benefits of masking arise from forcing the model to use different cues from the same video to predict the pseudolabel, resulting in more robust target domain recognition. As we saw in \cref{tab:stage_ablation}, the masked pre-training stage plays a key role in unlocking the benefits of masked self-training.

\input{tables/ab_celoss}

% Pseudolabeling strategy comparison
\noindent\textbf{Comparison of pseudolabeling strategies.}
In \cref{tab:pl_strategy}, we explore the use of alternative pseudolabeling strategies to the MatchOrConf \cite{zhangRethinkingRolePreTrained2023} scheme employed in \methodname.
The first 3 rows of the table assess variations of the masked consistency pseudolabeling scheme proposed in PACMAC \cite{prabhu2022adapting}, where target samples are selected if there is agreement between predictions of $k$ masked views. This criterion can also be combined with a constraint on the confidence exceeding threshold $T$. While we find the combination of these two constraints (ConsOrConf) to be effective for self-training, we find that MatchOrConf results in more consistent target accuracy improvements across domain shifts. We also consider combining MatchOrConf with a masked consistency constraint as a way to reduce cases where the student and teacher models agree on an incorrect prediction. This scheme, denoted as MatchAndConsOrConf, adds an additional constraint to the matching criterion by also requiring agreement between the student model predictions of $k$  masked views. We find that this approach underperforms compared to the vanilla MatchOrConf scheme due to low utilization of target domain samples. Our results also indicate that using both models together for pseudolableing outperforms using only zero-shot CLIP predictions.

\input{tables/ab_psudo}

% Masked Target During Self-Training
\noindent\textbf{Effect of source classification during self-training.}
In \cref{tab:pl_data_ablation}, we study the importance of including a source domain classification loss while performing collaborative self-training on masked target videos. Indeed, we find that source classification is a crucial ingredient in this stage, as excluding it destabilizes training and results in performance degradation rather than improvement. We suspect that the ground truth labels on source data help to combat the effects of inaccurate target pseudolabels, which may be prevalent especially at the start of training. 

\input{tables/ab_loss_st}

% Confidence threshold
\noindent\textbf{Choice of pseudolabeling confidence threshold.}
In \cref{tab:pl_gamma_sweep}, we compare adaptation performance for various values of confidence threshold in the MatchOrConf pseudolabeling scheme. We observe significant variation across domain shifts, but find that $\gamma=0.1$ leads to relatively consistent improvement. We leave for future work the development of a more principled approach to determining an appropriate confidence threshold for a given domain shift.

\input{tables/ab_match}

%% file: tables/ab_method.tex
\begin{table}[t]
\begin{center}
\resizebox{0.95\linewidth}{!}{
\begin{tabular}{cccccc}
\toprule
\multirow{3}{*}{\textbf{Method}} & \multirow{3}{*}{{PT}} & \multirow{3}{*}{{CST}} & \multicolumn{3}{c}{\textbf{Accuracy (\%)}} \\
& & & H$\rightarrow$A & A$\rightarrow$H  \\
\midrule
Source Only & \xmark & \xmark & 40.4 & 49.6  \\
\midrule
+ Target UMT Pre-Training & \cmark & \xmark & 43.8 & 51.7 \\
+ Collaborative Self-Training  & \xmark  & \cmark & 42.0 & 60.8  \\
\ourcolor \methodname & \ourcolor \cmark & \ourcolor \cmark & \ourcolor \textbf{48.0} & \ourcolor \textbf{67.9} \\
\bottomrule
\end{tabular}}
\end{center}
\vspace{-6mm}
\caption{Ablation of stages in \methodname on ARID$\leftrightarrow$HMDB from \textit{Daily-DA}. PT and CST denote UMT pre-training and collaborative self-training, respectively. }
\vspace{-2mm}
\label{tab:stage_ablation}
\end{table}

%% file: tables/ab_pretrain.tex
\begin{table}[t]
\begin{center}
\resizebox{0.65\linewidth}{!}{
\begin{tabular}{ccc} 
\toprule
\multirow{2}{*}{\textbf{UMT Pre-Training Data}} & \multicolumn{2}{c}{\textbf{Accuracy (\%)}} \\
& H$\rightarrow$A & A$\rightarrow$H \\
\midrule
None & 42.0 & 60.8 \\ 
Source Only & \textbf{48.3} & 60.0 \\
\ourcolor Target Only & \ourcolor \underline{48.0} & \ourcolor \textbf{67.9} \\
Source + Target & 45.2 & \underline{67.1} \\
\bottomrule
\end{tabular}}
\end{center}
\vspace{-6mm}
\caption{Comparison of data domains used during UMT pre-training. Accuracies are reported after the self-training stage. \textbf{Bold} indicates best accuracy and \underline{underline} indicates second best.}
\vspace{-2mm}
\label{tab:umt_data_ablation}
\end{table}

%% file: tables/ab_celoss.tex
\begin{table}[t]
\begin{center}
\resizebox{0.67\linewidth}{!}{
\begin{tabular}{ccc} 
\toprule
\multirow{2}{*}{\textbf{Data for Target CE Loss}} & \multicolumn{2}{c}{\textbf{Accuracy (\%)}} \\ 
& H$\rightarrow$A & A$\rightarrow$H \\
\midrule
Unmasked & 46.6 & 54.6 \\ 
\ourcolor Masked & \ourcolor \textbf{48.0} & \ourcolor \textbf{67.9} \\
\bottomrule
\end{tabular}}
\end{center}
\vspace{-6mm}
\caption{Effect of applying the target domain cross entropy loss on masked vs. unmasked target videos during the self-training stage.}
\vspace{-2mm}
\label{tab:pl_target_masking}
\end{table}

%% file: tables/ab_psudo.tex
\begin{table}[t]
\begin{center}
\resizebox{0.85\linewidth}{!}{
\begin{tabular}{lcc} 
\toprule
\multicolumn{1}{c}{\multirow{2}{*}{\textbf{Pseudolabeling Strategy}}} & \multicolumn{2}{c}{\textbf{Accuracy (\%)}} \\
& H$\rightarrow$A & A$\rightarrow$H \\
\midrule
Cons ($k=2$) & 39.8 & 57.5 \\
ConsOrConf ($k=2, T=0.5$) \cite{prabhu2022adapting} & \textbf{48.6} & 56.3 \\
ConsAndConf ($k=2, T=0.5$) & 40.1 & 57.1 \\
\color{black} Zero-Shot CLIP Only ($T=0.5$) & \color{black} 34.7 & \color{black} 51.3 \\
\ourcolor MatchOrConf ($\gamma=0.1$) \cite{zhangRethinkingRolePreTrained2023} & \ourcolor \underline{48.0} & \ourcolor \textbf{67.9} \\ 
MatchAndConsOrConf ($k=2, \gamma=0.1$) & 47.4 & \underline{60.8} \\
\bottomrule
\end{tabular}}
\end{center}
\vspace{-6mm}
\caption{Comparison of pseudolabeling strategies.}
\vspace{-2mm}
\label{tab:pl_strategy}
\end{table}

%% file: tables/ab_loss_st.tex
\begin{table}[t]
\begin{center}
\resizebox{0.72\linewidth}{!}{
\begin{tabular}{ccc} 
\toprule
\multirow{2}{*}{\textbf{Loss During Self-Training}} & \multicolumn{2}{c}{\textbf{Accuracy (\%)}} \\
& H$\rightarrow$A & A$\rightarrow$H \\
\midrule
 $\mathcal{L}_{\text{CE}}(m(\mathbf{x}^\target), \tilde{y})$ & 33.7 & 42.1 \\ 
\ourcolor $\mathcal{L}_{\text{CE}}(m(\mathbf{x}^\target), \tilde{y}) + \mathcal{L}_{\text{CE}}(\mathbf{x}^\source, {y^\source})$ & \ourcolor \textbf{48.0} & \ourcolor \textbf{67.9} \\
\bottomrule
\end{tabular}}
\end{center}
\vspace{-6mm}
\caption{Effect of including a source domain classification loss in addition to the target domain loss during the self-training stage.}
\vspace{-2mm}
\label{tab:pl_data_ablation}
\end{table}

%% file: tables/ab_match.tex
\begin{table}[t]
\begin{center}
\resizebox{0.6\linewidth}{!}{
\begin{tabular}{ccc} 
\toprule
\multirow{2}{*}{\textbf{Confidence Threshold}} & \multicolumn{2}{c}{\textbf{Accuracy (\%)}} \\
 & H$\rightarrow$A & A$\rightarrow$H \\
\midrule
\ourcolor 0.1 & \ourcolor \underline{48.0} & \ourcolor \textbf{67.9} \\ 
0.3 & 44.1 & 55.4 \\
0.5 & \textbf{51.9} & 64.6 \\
0.7 & 42.6 & \underline{65.8} \\
0.9 & 40.7 & 62.5 \\
\bottomrule
\end{tabular}}
\end{center}
\vspace{-6mm}
\caption{Effect of confidence threshold $\gamma$ in MatchOrConf pseudolabeling scheme from Equation \ref{eqn:matchORconf}.}
\vspace{-3mm}
\label{tab:pl_gamma_sweep}
\end{table}

%% file: sec/7_conclusions.tex
\section{Conclusions}
\label{sec:conclusions}

In this work, we presented \methodname, a three step process for unsupervised video domain adaptation. Our approach leverages a powerful spatial encoder to aid in the adaptation of a spatiotemporal network using unlabeled target domain videos--- making significant progress towards bridging the domain gaps in various VUDA  datasets. The advancements demonstrated by \methodname underscore the untapped potential of masked modeling techniques for video domain adaptation, which we hope will spur further research in this area. \smallskip
\vspace{-2mm}

\noindent \small {\textbf{Acknowledgements.} This research was sponsored by the Army Research Laboratory under Cooperative Agreement W911NF-21-2-0211. 
The views and conclusions contained in this document are those of the authors and should not be interpreted as representing the official policies, either expressed or implied, of the Army Research Office or the U.S. Government. 
The U.S. Government is authorized to reproduce and distribute reprints for Government purposes notwithstanding any copyright notation herein.}

%% file: sec/X_suppl.tex
\clearpage
\setcounter{page}{1}
\maketitlesupplementary

\section{Additional Training Details}
\label{sec:imp_deets_supp}

In Tables \ref{tab:stage1_deets}, \ref{tab:stage2_deets} and \ref{tab:stage3_deets} we provide a more detailed account of the training configurations used in each of the three stages of \methodname. The ViT-B network we use has 87M parameters, all of which are trained in each of the 3 stages of \methodname. Stage 2, which is analogous to standard fine-tuning, took roughly 6 hours for K$\rightarrow$S (the largest task we evaluate on) whereas our additional stages, Stage 1 and Stage 3, took 1.5 hours and 6 hours respectively on 4 NVIDIA A5000 GPUs.

\begin{table}[h]
\centering
\resizebox{0.76\linewidth}{!}{\begin{tabular}{c|c}
\toprule
\textbf{Setting}        & \textbf{Value}            \\ \midrule

Learning Rate Schedule & Cosine \\
Base Learning Rate & 1.5e-4 \\
Batch Size & 256 \\
Warmup Epochs (Linear) & 10 \\
Total Epochs & 50 \\
Optimizer & AdamW \cite{loshchilov2017decoupled} \\
Optimizer Betas & $\beta_1=0.9$, $\beta_2=0.95$ \\
Weight Decay & 0.05 \\
Drop Path \cite{huang2016deep} & 0.1  \\
Horizontal Flip & Yes \\
Random Resize Scales & [0.66, 0.75, 0.875, 1] \\
Masking Ratio & 0.8 \\
\bottomrule
\end{tabular}}
\caption{Training configuration for UMT pre-training stage of \methodname (Stage 1).}
\label{tab:stage1_deets}
\end{table}

\begin{table}[h]
\centering
\resizebox{0.76\linewidth}{!}{\begin{tabular}{c|c}
\toprule
\textbf{Setting}        & \textbf{Value}            \\ \midrule

Learning Rate Schedule & Cosine \\
Base Learning Rate & 2.5e-5 \\
Batch Size & 28 \\
Warmup Iterations (Linear) &  4,000 \\
Total Iterations &  20,000 \\
Optimizer & AdamW \cite{loshchilov2017decoupled} \\
Optimizer Betas & $\beta_1=0.9$, $\beta_2=0.999$ \\
Weight Decay & 0.05 \\
Drop Path \cite{huang2016deep} & 0.1  \\
Layer-Wise LR Decay & 0.65 \\
Random Erase & 0.25 \\
RandAug \cite{cubuk2019randaugment} & $M=7$, $N=4$ \\
\bottomrule
\end{tabular}}
\caption{Training configuration for source domain fine-tuning stage of \methodname (Stage 2).}
\label{tab:stage2_deets}
\end{table}

\begin{table}[ht]
\centering
\resizebox{0.9\linewidth}{!}{\begin{tabular}{c|c|c}
\toprule
\textbf{Data} & \textbf{Setting}        & \textbf{Value}            \\ \midrule
\scriptsize \multirow{10}{*}{\rotatebox{0}{Common}}
& LR Schedule & Cosine \\
& Learning Rate & 1e-5 \\
& Warmup Iterations & 4,000 \\
& Total Iterations &  20,000 \\
& Optimizer & AdamW \cite{loshchilov2017decoupled} \\
& Optimizer Betas & $\beta_1,\beta_2=0.9,0.95$\\
& Weight Decay & 0.05 \\
& Drop Path \cite{huang2016deep} & 0.1  \\
& Layer-Wise LR Decay & 0.75 \\
& MatchOrConf Threshold ({$\gamma$}) & 0.1 \\
\midrule
\scriptsize \multirow{3}{*}{\rotatebox{0}{Source}}
& Batch Size & 20 \\
& Random Erase & 0.25 \\
& RandAug \cite{cubuk2019randaugment} & $M=7$, $N=4$ \\
\midrule
\scriptsize \multirow{2}{*}{\rotatebox{0}{Target}}
& Batch Size & 20 \\
& Data Transform & CenterCrop \\
\midrule
\scriptsize \multirow{2}{*}{\rotatebox{0}{Masked Target}}
& Batch Size & 20 \\
% & RandAug \cite{cubuk2019randaugment} & $M=3$, $N=2$ \\
& Loss Coeff. ($\lambda$) & 1 \\
\bottomrule
\end{tabular}}
\caption{Training configuration for collaborative self-training stage of \methodname (Stage 3). In order to enhance pseudolabel accuracy, unmasked target domain videos do not undergo data augmentation.}
\label{tab:stage3_deets}
\end{table}

\begin{table*}[th]
\begin{center}
\small
\resizebox{1.0\linewidth}{!}{
\begin{tabular}{c|c|ccc|ccc|ccc|ccc|c}
\toprule
\multirow{2}{*}{\textbf{Initialization}} & \multirow{2}{*}{\textbf{Method}} & \multicolumn{12}{c}{\textbf{Target Domain Accuracy (Top-1\%)}} \\
& & H→A & M→A & K→A & A→H & M→H & K→H & H→M & A→M & K→M & M→K & H→K & A→K &  {Avg.} \\
\midrule
\scriptsize \multirow{2}{*}{\rotatebox{0}{UMT K710}}
& \socolor Source Only & \socolor 40.4 & \socolor {52.1} & \socolor 36.5 & \socolor 49.6 & \socolor 68.3 & \socolor 57.9 & \socolor 41.5 & \socolor 36.3 & \socolor 43.3 & \socolor 79.3 & \socolor 48.0 & \socolor 41.7 & \socolor 49.6 \\
& \tocolor Target Only & \tocolor 68.5 & \tocolor 68.5 & \tocolor 68.5 & \tocolor 84.6 & \tocolor 84.6 & \tocolor 84.6 & \tocolor 73.0 & \tocolor 73.0 & \tocolor 73.0 & \tocolor 98.3 & \tocolor 98.3 & \tocolor 98.3 & \tocolor 81.1 \\
\midrule
\scriptsize \multirow{2}{*}{\rotatebox{0}{+ K400 Sup.}}
& \socolor Source Only & \socolor 57.2 & \socolor 74.7 & \socolor 54.0 & \socolor 70.8 & \socolor 71.7 & \socolor 58.3 & \socolor 57.8 & \socolor 51.8 & \socolor 44.3 & \socolor 91.2 & \socolor 65.4 & \socolor 61.7 & \socolor 63.2 \\
& \tocolor Target Only & \tocolor 81.5 & \tocolor 81.5 & \tocolor 81.5 & \tocolor 90.4 & \tocolor 90.4 & \tocolor 90.4 & \tocolor 80.0 & \tocolor 80.0 & \tocolor 80.0 & \tocolor 99.6 & \tocolor 99.6 & \tocolor 99.6 & \tocolor 87.9 \\
\bottomrule
\end{tabular}}
\end{center}
\vspace{-4mm}
\caption{\textit{Daily-DA} source only and target only baselines using self-supervised vs. supervised Kinetics pre-trained weights for initialization. ``UMT K710" denotes self-supervised UMT pre-training on Kinetics-710, while ``+ K400 Sup." denotes additional supervised fine-tuning on Kinetics-400.}
\label{tab:k400_daily-da}
\end{table*}

\begin{table*}[th]
\small
\centering
\resizebox{0.6\linewidth}{!}{
\begin{tabular}{c|c|cc|cc|cc|c}
\toprule
\multirow{2}{*}{\textbf{Initialization}} & \multirow{2}{*}{\textbf{Method}} & \multicolumn{5}{c}{\textbf{Target Domain Accuracy (Top-1 \%)}} \\
& & U→S & K→S & S→U & K→U & U→K & S→K & {Avg.} \\
\toprule
\scriptsize \multirow{2}{*}{\rotatebox{0}{UMT K710}}
& \socolor Source Only & \socolor 67.6 & \socolor 86.8 & \socolor 97.9 & \socolor 98.9 & \socolor 79.9 & \socolor 89.1 & \socolor 86.7 \\ 
& \tocolor Target Only & \tocolor 97.9 & \tocolor 97.9 & \tocolor 98.8 & \tocolor 98.8 & \tocolor 99.9 & \tocolor 99.9 & \tocolor 98.9 \\ 
\midrule
\scriptsize \multirow{2}{*}{\rotatebox{0}{+ K400 Sup.}}
& \socolor Source Only & \socolor 83.1 & \socolor 86.9 & \socolor 99.3 & \socolor 99.9 & \socolor 95.9 & \socolor 95.7 & \socolor 93.5 \\
& \tocolor Target Only & \tocolor 96.4 & \tocolor 96.4 & \tocolor 99.5 & \tocolor 99.5 & \tocolor 98.7 & \tocolor 98.7 & \tocolor 98.2 \\ 

\bottomrule
\end{tabular}}
\caption{\textit{Sports-DA} source only and target only baselines using self-supervised vs. supervised pre-trained weights for initialization. ``UMT K710" denotes self-supervised UMT pre-training on Kinetics-710, while ``+ K400 Sup." denotes additional supervised fine-tuning on Kinetics-400.}
\label{tab:k400_sports-da}
\end{table*}

\section{Masked Consistency Implementation}
\label{sec:pacmac_supp}

In \cref{tab:pl_strategy}, we investigated alternative pseudolabeling strategies from the MatchOrConf scheme that we employ in \methodname. Rows 1, 2, 3 and 5 include a masked consistency constraint in the style of PACMAC \cite{prabhu2022adapting}. Here, we provide more detail on the implementation of this masked consistency constraint. 

A video is said to satisfy the masked consistency constraint if the class prediction of $f_\student$ on the full (\ie unmasked) video is consistent with the class predictions of each of the $k$ masked versions. Like \cite{prabhu2022adapting}, we form the masked views using an attention-guided, greedy round-robin assignment process. Because $f_\student$ uses mean pooling instead of a CLS token for classification, we use the attention map of the CLIP teacher image encoder to create a mask for each video frame, with a masking ratio of $r=0.8$ (identical to the masking process used in the UMT pre-training stage of \methodname). The result is $k$ disjoint masks for each video frame, which are then applied to the video to create the $k$ masked views for consistency assessment.

\section{Supervised Kinetics-400 Initialization}
\label{sec:k400_init_results_supp}

In \cref{sec:initialization}, we discussed the motivation behind initializing our student network from self-supervised UMT pre-training on Kinetics-710 rather than the supervised Kinetics-400 initialization that has become common in the video DA literature. In Tables \ref{tab:k400_daily-da} and \ref{tab:k400_sports-da} we provide a comparison of source only and target only baselines on \textit{Daily-DA} and \textit{Sports-DA} using both initializations. As expected, we observe significantly higher baseline accuracies when using a supervised Kinetics initialization.

\section{Zero-Shot Classification with CLIP}
\label{sec:clip_supp}
The process we use to perform zero-shot image classification using CLIP \cite{radfordLearningTransferableVisual2021a} is described in \cref{sec:imp_deets}. In \cref{tab:class-names}, we provide the exact class names used to form the inputs to the CLIP text encoder. Classification is performed using a single template: ``\textit{A video of a person} \{class\}."

\begin{table}[!h]
\centering
\resizebox{0.82\linewidth}{!}{\begin{tabular}{c|c|c}
\toprule
\textit{Daily-DA}             & \textit{Sports-DA}               & \textit{UCF$\leftrightarrow$HMDB}\(_{full}\)           \\ \midrule
drink                         & archery                          & climb                              \\
jump                          & baseball                         & fencing                            \\
pick                          & basketball                       & golf                               \\
pour                          & biking                           & soccer                             \\
push                          & bowling                          & pullup                             \\
run                           & swimming                         & boxing                             \\
walk                          & diving                           & pushup                             \\
wave                          & fencing                          & riding bike                        \\
                              & field hockey                     & horse riding                       \\
                              & gymnastics                       & basketball                         \\
                              & golf                             & archery                            \\
                              & horse riding                     & walking                            \\
                              & kayaking                         &                                     \\
                              & rock climbing                    &                                     \\
                              & climbing rope                    &                                     \\
                              & skateboarding                    &                                     \\
                              & skiing                           &                                     \\
                              & sumo wrestling                    &                                     \\
                              & surfing                          &                                     \\
                              & tai chi                          &                                     \\
                              & tennis                           &                                     \\
                              & trampoline jumping               &                                     \\
                              & volleyball                       &                                     \\ \bottomrule

\end{tabular}}
\caption{Class names used in zero-shot CLIP classification for each VUDA benchmark.}
\label{tab:class-names}
\end{table}